\documentclass[11pt]{article}
\usepackage{coling2020,times,hyperref,natbib,booktabs,microtype,xcolor,float,
    graphicx,mdwlist,soul,linguex}

\colingfinalcopy
\definecolor{darkblue}{rgb}{0, 0, 0.5}  
\hypersetup{colorlinks=true, citecolor=darkblue,
    linkcolor=darkblue, urlcolor=darkblue}
  
\title{A Benchmark of Rule-Based and Neural \\
    Coreference Resolution in Dutch Novels and News}

\author{Corb\`en Poot \\
  CLCG, University of Groningen \\
  {\tt c.poot@student.rug.nl} \\\And
  Andreas van Cranenburgh \\
  CLCG, University of Groningen \\
  {\tt a.w.van.cranenburgh@rug.nl} \\}

\date{}

\begin{document}
\maketitle
\begin{abstract}
    We evaluate a rule-based \citep{lee2013deterministic}
    and neural \citep{lee2018higher} coreference system on Dutch datasets of
    two domains: literary novels and news/Wikipedia text.
    The results provide insight into the relative strengths of data-driven and
    knowledge-driven systems, as well as the influence of domain, document
    length, and annotation schemes.
    The neural system performs best on news/Wikipedia text,
    while the rule-based system performs best on literature.
    The neural system shows weaknesses with limited training data and long
    documents, while the rule-based system is affected by annotation
    differences. The code and models used in this paper are available at
    \url{https://github.com/andreasvc/crac2020}
\end{abstract}

\section{Introduction}
In recent years, the best results for coreference resolution
of English have been obtained with
end-to-end neural models~\citep{lee2017neural, lee2018higher, joshi2019bert,
    joshi2020spanbert, wu2020corefqa}.
However for Dutch, the existing systems are still using either a 
rule-based~\citep{vandergoot2015, vancranenburgh2019coref} or a machine
learning approach~\citep{hendrickx2008coreference, declercq2011cross}.
The rule-based system dutchcoref~\citep{vancranenburgh2019coref}
outperformed previous systems on two existing datasets
and also presented a corpus and evaluation of literary novels (RiddleCoref).

In this paper we compare this rule-based system
to an end-to-end neural coreference
resolution system: e2e-Dutch.
This system is a variant of \citet{lee2018higher} with BERT token
representations.
We evaluate and compare the performance of e2e-Dutch to dutchcoref on two
different datasets:
(1) the SoNaR-1 corpus \citep{schuurman2010interacting},
a genre-balanced corpus of 1 million words,
and
(2) the RiddleCoref corpus of contemporary novels \citep{vancranenburgh2019coref}.
This provides insights into
(1) the relative strengths of a neural system versus a rule-based system for
Dutch coreference, and (2) the effect of domain differences (news/Wikipedia
versus literature).

The two datasets we consider
vary greatly in terms of overall size
and length of the individual documents;
the training subset of RiddleCoref contains only 23 documents
(novel fragments)
compared to 581 documents for SoNaR-1.
However, the average number of sentences per document is higher
for RiddleCoref than for SoNaR-1 (295.78 vs. 64.28 respectively). 
We also conduct an error analysis for both of the systems
to examine the types of errors that the systems make. 

\section{Related work}
The main differences between traditional and neural approaches can be
summarized as follows:

\begin{itemize*}
    \item 
        Rule-based systems are knowledge-intensive; machine learning systems
        are data-driven but require feature engineering;
        end-to-end neural systems only require sufficient training data
        and hyperparameter tuning
        to perform well.
    \item 
        Rule-based and machine learning coreference systems
        rely on features from syntactic parses and named-entities
        provided by an NLP pipeline
        whereas neural systems rely on distributed representations;
        end-to-end systems do not require any other features.
    \item 
        The rule-based system by \citet{lee2013deterministic}
        is entity-based and exploits global features,
        while end-to-end systems
        such as \citet{lee2017neural} rank mentions
        and make greedy decisions based on local features.
        Although \citet{lee2018higher} does approximate
        higher-order inference, their model does not build
        representations of entities.
\end{itemize*}

The rest of this section discusses the current best systems
for Dutch and English.

\subsection{Dutch coreference resolution}

The largest dataset available for Dutch coreference resolution is the SoNaR-1
dataset \citep{schuurman2010interacting} which consists of 1 million words
annotated for coreference. This corpus was a continuation of the Corea project
\citep{bouma2007corea, hendrickx2008coreference, hendrickx2008semantic}.
\citet{declercq2011cross} present a cross-domain coreference resolution study
conducted on this corpus. They use a mention-pair system, which was originally
developed with the KNACK-2002 corpus and then further improved in the Corea
project, and observe that the influence of domain and training size is large,
thus underlining the importance of this large
and genre-balanced SoNaR-1 dataset.

The current best coreference resolution system for Dutch is called
``dutchcoref" \citep{vancranenburgh2019coref} and is based on the rule-based
Stanford system \citep{lee2011multipass, lee2013deterministic}. This system
improved on the systems in the SemEval-2010 shared
task~\citep{semeval2010coref} and a previous implementation of the Stanford
system for Dutch \citep[GroRef; ][]{vandergoot2015}.
The main focus of \citet{vancranenburgh2019coref} was evaluating coreference on
literary texts, for which a corpus and evaluation is presented.
Most coreference resolution systems are evaluated using newswire texts, but a
domain such as literary text presents its own
challenges \citep{bamman2017natural};
for example, novels are longer than news articles,
and novels can therefore contain longer coreference chains.

\subsection{English Coreference resolution}

\begin{table}[t]\centering\footnotesize
\begin{tabular}{lr}\toprule
System                                          & CoNLL\\ \midrule
Rule-based \citep{lee2011multipass}             & 58.3 \\
Perceptron \citep{fernandes2012latent}          & 58.7 \\
Hybrid: rules + ML \citep{lee2017scaffolding}   & 63.2 \\
Embeddings \citep{wiseman2015learning}          & 63.4 \\
+ RL \citep{clark2016deep}                      & 65.3 \\
+ Entity embeddings \citep{clark2016improving}  & 65.7 \\
\bottomrule\end{tabular}~~~~~~~~~\begin{tabular}{lr}\toprule
System                                          & CoNLL\\ \midrule
End-to-end \citep{lee2017neural}                & 68.8 \\
Higher-order + CTF + ELMo \citep{lee2018higher} & 73.0 \\
Finetuning BERT base \citep{joshi2019bert}      & 73.9 \\
Finetuning BERT large \citep{joshi2019bert}     & 76.9 \\
Pretraining SpanBERT \citep{joshi2019bert}      & 79.6 \\
SpanBERT + QA \citep{wu2020corefqa}             & 83.1 \\ \bottomrule
\end{tabular}
\caption{English coreference scores on the OntoNotes CoNLL 2012 shared task
    dataset. ML: Machine Learning, RL: Reinforcement Learning, CTF:
    Coarse-to-Fine, QA: Question Answering.}\label{tblenglishsota}
\end{table}

The main benchmark for English is the CoNLL 2012 shared task \citep{conll2012}.
\autoref{tblenglishsota} reports a timeline of results for this task,
which shows the dramatic improvements brought by neural networks, especially
the end-to-end systems on the right. Neural coreference systems improved on
previous work but were still relying on mention detection rules, syntactic
parsers, and heavy feature engineering (\autoref{tblenglishsota}, left).
They were outperformed by the first end-to-end coreference
resolution system by \citet{lee2017neural}. This system looks at all 
the spans (expressions) in a text, up to a maximum length, and then
uses a span-ranking model that decides for each span which previous
spans are good antecedents, if any. The spans themselves are
represented by word embeddings.

Although the models by \citet{clark2016deep} and \citet{lee2017neural}
are computationally efficient and scalable to long documents, 
they are heavily relying on first order models where they are only
scoring pairs of mentions. Because they make independent
decisions regarding coreference links, they might make predictions
which are locally consistent but globally inconsistent
\citep{lee2018higher}. \citet{lee2018higher} introduce
an approximation of higher-order inference, which uses the span-ranking
architecture from \citet{lee2017neural} described above in an 
iterative fashion,
and also propose a coarse-to-fine approach to 
lower the computational cost of this iterative higher-order
approximation. Further improvements over \citet{lee2017neural} 
were obtained through the use of deep contextualized ELMo
\citep{peters2018deep} word embeddings.
The current state-of-the-art scores are even higher
by using BERT finetuning \citep{joshi2019bert,joshi2020spanbert,wu2020corefqa}
However, this paper focuses on the model by \citet{lee2018higher}.

\citet{bamman2019coref} present coreference results on English literature with
an end-to-end model comparable to the one used in this paper, except for using
a separate mention detection step. However, their dataset consist of a larger
number of shorter novel fragments (2000 words). They report a CoNLL score of
68.1 on the novel fragments.

\begin{table}[t]\centering\footnotesize
\begin{tabular}{@{}lrrrrrr@{}}\toprule
                  & \multicolumn{3}{c}{RiddleCoref}  & \multicolumn{3}{c}{SoNaR-1} \\ \cmidrule(lr){2-4} \cmidrule(lr){5-7}
                  & Train   & Dev     & Test    & Train   & Dev     & Test    \\ \midrule
documents         & 23      & 5       & 5       & 581     & 135     & 145     \\
sentences         & 6803    & 1525    & 1536    & 37,346  & 10,585  & 11,671  \\
tokens            & 105,517 & 28,042  & 28,054  & 635,191 & 171,293 & 197,392 \\
sents per doc     & 295.78  & 305     & 307.2   & 64.28   & 78.41   & 80.49   \\
avg sent len      & 15.51   & 18.39   & 18.26   & 17      & 16.18   & 16.91   \\
mentions          & 25,194  & 6584    & 6869    & 182,311 & 50,472  & 57,172  \\
entities          & 9041    & 2643    & 3008    & 128,142 & 37,057  & 39,904  \\
mentions/tokens   & 0.24    & 0.23    & 0.24    & 0.29    & 0.29    & 0.29    \\
mentions/entities & 2.79    & 2.49    & 2.28    & 1.42    & 1.36    & 1.43    \\
entities/tokens   & 0.09    & 0.09    & 0.11    & 0.20    & 0.22    & 0.20    \\
\% pronouns       & 40.4    & 35.7    & 38.1    & 11.6    & 11.3    & 11.0    \\    
\% nominal        & 47.0    & 49.4    & 52.8    & 70.8    & 70.4    & 71.9    \\
\% names          & 12.6    & 14.9    & 9.1     & 17.6    & 18.3    & 17.1    \\
\bottomrule
\end{tabular}
\caption{Dataset statistics}\label{tbldata}
\end{table}

\section{Coreference corpora}\label{seccorpora}

In this paper we consider entity coreference
and focus on the relations of identity and predication.
The rest of this section describes the two Dutch corpora we use.

\subsection{SoNaR-1: news and Wikipedia text}
The SoNaR-1 corpus \citep{schuurman2010interacting} contains about
1 million words of Dutch text from various genres, predominantly news
and Wikipedia text. Coreference was annotated from scratch
(i.e., annotation did not proceed by correcting the output of a system),
based on automatically extracted markables.
The markables include singleton mentions
but also non-referring expressions such as pleonastic pronouns.
The annotation was not corrected by a second annotator.
\citet{hendrickx2008semantic} estimated the inter-annotator agreement
of a different corpus with the same annotation scheme and obtained
a MUC score of 76 \% for identity relations (which form the majority).

We have created a genre-balanced train/dev/test split for SoNaR-1
of 70/15/15.
The documents are from a range of different genres
and we therefore ensure that the subsets
are a stratified sample in terms of genres,
to avoid distribution shifts between the train and test set.\footnote{%
    Cf.~\url{https://gist.github.com/CorbenPoot/ee1c97209cb9c5fc50f9528c7fdcdc93}}.

We convert the SoNaR-1 coreference annotations
from MMAX2 format into the CoNLL-2012 format.
Since dutchcoref requires parse trees as input,
we use the manually corrected Lassy Small treebank~\citep{noord2006syntactic,lassy},
which is a superset of the SoNaR-1 corpus.\footnote{%
    We could also evaluate with predicted parses from the Alpino parser, but
    components of the Alpino parser have been trained on subsets of Lassy
    Small, so predicted parses of Lassy Small are not representative of
    Alpino's heldout performance.}
We align the Lassy Small trees at the sentence and token level to the SoNaR-1
coreference annotations, since there are some differences in tokenization and
sentence order.\footnote{%
    The conversion script is part of
    \url{https://github.com/andreasvc/dutchcoref/}}
We also add gold standard NER annotations from SoNaR-1.
The manually corrected trees lack some additional features produced
by the Alpino parser~\citep{noord2006alpino} which are needed by dutchcoref;
we merge these predicted features into the gold standard trees.

\subsection{RiddleCoref: contemporary novels}
The RiddleCoref corpus consists of
contemporary Dutch novels
(both translated and originally Dutch),
and was presented in \citet{vancranenburgh2019coref}.
The corpus is a subset of the Riddle of Literary Quality corpus of
401 bestselling novels~\citep{koolen2020survey}.
This dataset was annotated by
correcting the output of dutchcoref. Most novels in the
dataset were corrected by two annotators, with the second performing
another round of correction after the first.
In this dataset,
mentions include singletons and are manually corrected; i.e., only expressions
that refer to a person or object are annotated as mentions.
Besides this difference, relative clauses and discontinuous constituents have
different boundaries (minimal spans).

The system by \citet{vancranenburgh2019coref} is a rule-based system that
does not require a training data,
and therefore the dev/test split used in this
paper is not suitable for a supervised system. To avoid this issue,
we create a new train/dev/test split which reserves 70\% for
training data. We also evaluate dutchcoref on this new split.
The new dev and test sets have no overlap with the original development set
on which the rules of dutchcoref were tuned.

No gold standard parse trees are available for the novels.
Instead, we use automatically predicted parses from the Alpino parser~\citep{noord2006alpino}.

\subsection{Dataset statistics}
\autoref{tbldata} shows statistics of the two datasets and their respective
splits. The documents in RiddleCoref are almost four times as long as those in
SoNaR-1, and this is reflected in a higher number of mentions per entity, while
SoNaR-1 has a higher density of entities to tokens. We also see a difference
due to the more selective, manual annotation of mentions:
almost 30\% of SoNaR-1 tokens are part of a mention, compared to
less than 25\% for RiddleCoref. Finally, we see large differences in the
proportion of pronouns, nominals and names, due to the genre difference.

\section{Coreference systems}\label{secsystems}
We now describe the two coreference systems,
dutchcoref and e2e-Dutch,
which we evaluate on the coreference corpora
described in the previous section.

\subsection{Rule-based: dutchcoref}
The dutchcoref system\footnote{%
    \url{https://github.com/andreasvc/dutchcoref}}~\citep{vancranenburgh2019coref}
is an implementation of the rule-based coreference system by
\citet{lee2011multipass,lee2013deterministic}. The input to the system consists
of Alpino parse trees \citep{noord2006alpino}, which include named entities.
The system infers information about speakers and addressees of direct speech
using heuristic rules. This information is used for coreference decisions. Note
that this information is not given as part of the input.

We have made some improvements to the rules of this system in order to
make it more compatible with the SoNaR-1 annotations; this was however
based only on the output of a single document in the development set, as well
as on the original, RiddleCoref development set on which dutchcoref was
developed. When evaluating on SoNaR-1, we apply rules to filter links and
mentions from the output to adapt to the annotation scheme of this dataset.

\begin{figure}\centering
    \includegraphics[width=0.8\linewidth]{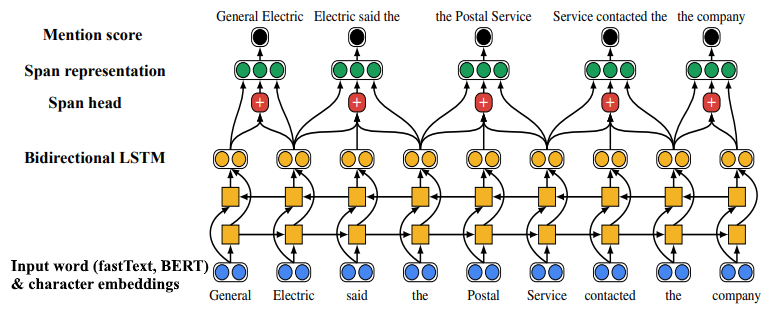}
    \caption{Overview of the first step of the end-to-end model
    in which the embedding representations and mention scores are computed.
    The model considers all possible spans up to a maximum width but only a
    small subset is shown here. Figure adapted from \citet{lee2017neural}.
        }\label{figmodel}
\end{figure}

\subsection{End-to-end, neural: e2e-Dutch}
The e2e-Dutch system\footnote{%
    The e2e-Dutch system is being developed
    as part of the Filter Bubble project at the VU and eScience center.
    The specific commit we used is \url{https://github.com/Filter-Bubble/e2e-Dutch/tree/056dcf7d3d711a3c7b8cda241a16cdd76158a823}
    }
is fully end-to-end in the sense that it is trained only on the token and
coreference column of the CoNLL files of the dataset, without using any
metadata. Our data does not contain speaker information which is used by models
trained on the OntoNotes dataset \citep{hovy2006ontonotes}.
In addition, models trained on OntoNotes use genre information; while our data
does have genre metadata, we have not experimented with using this feature. For
English, such information provides additional improvement in
scores \citep{lee2017neural}.

The model that e2e-Dutch is based on \citep{lee2018higher}  uses a combination
of character n-gram embeddings, non-contextual word embeddings~\citep[GloVe;
][]{pennington2014glove} and contextualized word embeddings~\citep[ELMo;
][]{peters2018deep}. These embeddings are concatenated and fed into a
bidirectional LSTM.
Span heads are approximated using an attention mechanism;
while this step is intended to approximate syntactic heads,
it does not rely on parse tree information.
\autoref{figmodel} shows an overview of the model.
e2e-Dutch adapts this architecture by adding support for singletons; i.e.,
during mention detection, each span is classified as not a mention, a
singleton, or a coreferent mention.

Character n-gram embeddings are extracted by iterating over the data and
feeding the character n-grams to a Convolutional Neural Network (CNN) which
then represents these n-grams as learned 8-dimensional embeddings. 
The GloVe embeddings were replaced with fastText\footnote{%
    We use Fasttext common crawl embeddings,
    \url{https://fasttext.cc/docs/en/crawl-vectors.html}}
embeddings~\citep{grave2018learning}. We also trained fastText embeddings on
our own datasets but saw a performance decrease; we therefore stick with
pre-trained embeddings.
Lastly, the ELMo embeddings were replaced by BERT \citep{devlin2019bert} token
embeddings, since BERT tends to outperform ELMo \citep{devlin2019bert}
and because there is a pretrained, monolingual Dutch BERT model available whose
pretraining data includes novels~\citep[BERTje; ][]{devries2019bertje}.
However, there is no overlap between the 7000+ novels that BERTje is trained on
and the RiddleCoref corpus. Whenever there is a mismatch between the subtokens
of BERT and the tokens in the coreference data, the model takes the average of
the BERT subtoken embeddings as token representation.
The last BERT layer is used for the token representation;
however, recent research has showed that layer 9 actually
performs best for Dutch coreference~\citep{vries2020special}.
Note also that we do not finetune BERT for this task,
contrary to \citet{joshi2019bert}; this is left for future work.

We use some different hyperparameters compared to \citet{lee2018higher}.
Our model only considers up to 30 antecedents per span
instead of 50;
this only leads to marginally worse performance,
a 0.03 decrease in the LEA F1-score,
while reducing the computational cost substantially.
During training, each document is randomly truncated at 30 sentences,
but different random parts are selected at each epoch.
We have experimented with higher values for this parameter with RiddleCoref,
but only obtained marginal improvements (0.01 difference),
and did not pursue this further.
The top span ratio controls the number of mentions that are considered and
determines the precision/recall tradeoff for mentions. We experimented with
tuning this parameter, but settled on the default of 0.4.
Mentions up to 50 tokens long are considered.

During training, the model is evaluated every 1500 epochs (2500 for SoNaR-1).
If the CoNLL score on the development set does not increase after three rounds,
training is stopped.

\section{Evaluation}
Before presenting our main benchmark results,
we discuss the issue of coreference evaluation metrics.

\subsection{Metrics}

The challenge with evaluating coreference resolution lies in the fact
that it involves several levels: mentions, links and entities.
Results can be correct on one level and incorrect on another,
and the levels interact.
One of the most important factors in coreference performance is
the performance of mention detection, since an incorrect or missed mention
can lead to a large number of missed coreference links
(especially for a long coreference chain).
We therefore report mention scores.
It turns out that mention performance
also has a large influence on coreference evaluation metrics~\citep{moosavi2016coreference}.
We will use two coreference metrics. The CoNLL score~\citep{conll2011}
is the standard benchmark,
but it does not have a precision and recall score,
and the MUC, $B^3$, and CEAFe metrics
on which it is based have their own flaws.
Therefore we will also look at the LEA metric \citep{moosavi2016coreference}.
LEA gives more weight to larger entities, so that mistakes on
more important chains have more effect on the score than mistakes
on smaller entities.

Unless otherwise noted, all our results include singletons.
Evaluating with and without singletons will affect all of the scores,
and the two datasets differ in the way they annotated singletons.
Singletons inflate coreference scores due to the mention identification effect.
Since most mentions are easy to identify based on form,
singletons reduce the informativeness of the coreference score.
SoNaR-1 includes automatically extracted markables
instead of manually annotated mentions, as in RiddleCoref.
The automatically extracted markables are more numerous
and easier to identify (they were extracted based on syntax)
than manually annotated mentions that are restricted to potentially
referring expressions (a semantic distinction).
One possibility to rule out the mention identification effect completely
is to present the systems with gold mentions. However, this still leaves the
singleton-effect. If singletons are included, the system will not know which
of the gold mentions are singletons, and this can lead to incorrect coreference
links. A dataset with more singletons (such as SoNaR-1) will thus have more
potential for incorrect coreference links (precision errors). If singleton
mentions are excluded from the set of gold mentions, it is  given that all
mentions are coreferent. The system should then use this information
and force every mention to have at least one link.
However, this requires re-training or re-designing the coreference system,
and does not allow us to do a realistic end-to-end coreference evaluation.
We are therefore stuck with the complications that come with
combining mention identification and coreference resolution.

\begin{table}[t]\centering
\begin{tabular}{llrrrrrrr}\toprule
System     & dataset      & \multicolumn{3}{c}{Mentions} & \multicolumn{3}{c}{LEA}  & CoNLL \\
           \cmidrule(lr){3-5} \cmidrule(lr){6-8}
           &              & R & P & F1                   & R & P & F1               &        \\ \midrule
dutchcoref & RiddleCoref, dev  & 86.85 & 85.84 & 86.34 & 49.18 & 58.03 & \bf 53.24 & \bf 65.91 \\
e2e-Dutch  & RiddleCoref, dev  & 83.12 & 87.65 & 85.33 & 48.37 & 50.99 & 49.65 & 64.81 \\ \midrule
dutchcoref & RiddleCoref, test & 87.65 & 90.80 & 89.20 & 50.83 & 64.78 & \bf 56.97 & \bf 69.86 \\
e2e-Dutch  & RiddleCoref, test & 81.95 & 89.00 & 85.33 & 44.82 & 50.48 & 47.48 & 63.55 \\ \midrule
dutchcoref & SoNaR-1, dev   & 64.88 & 86.78 & 74.25 & 37.98 & 52.23 & 43.98 & 55.45 \\
e2e-Dutch  & SoNaR-1, dev   & 90.24 & 88.09 & 89.16  & 65.02 & 65.55 & \bf 65.29 & \bf 71.53 \\ \midrule
dutchcoref & SoNaR-1, test  & 65.32 & 85.94 & 74.22 & 37.87 & 52.55 & 44.02 & 55.91 \\
e2e-Dutch  & SoNaR-1, test  & 88.96 & 86.81 & 87.87 & 60.67 & 62.48 & \bf 61.56 & \bf 68.45 \\ \bottomrule
\end{tabular}
\caption{Coreference results (predicted mentions, including singletons).}\label{tblresults}
\end{table}

\begin{figure}\centering
    \includegraphics[width=\linewidth]{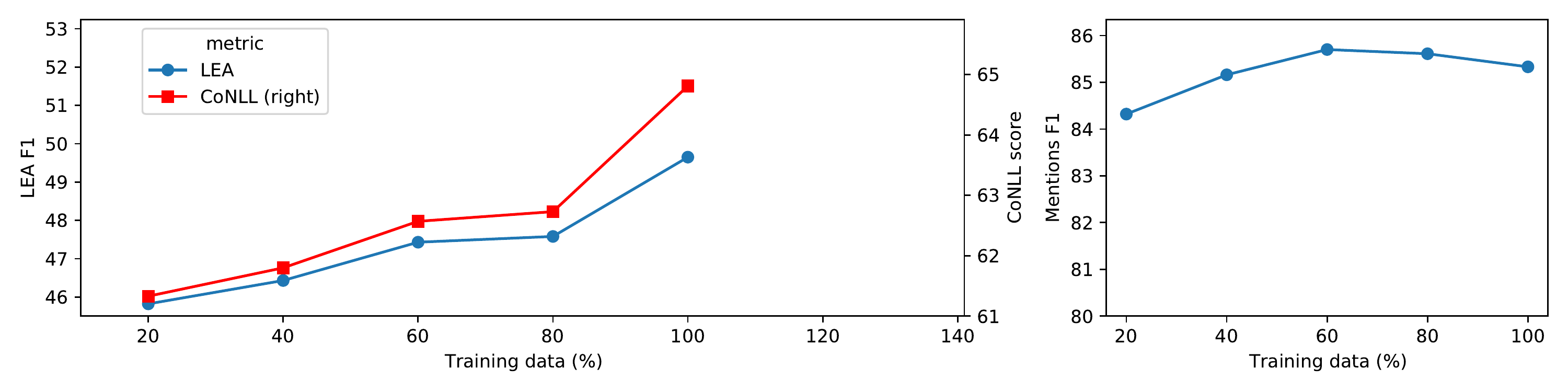}
    \caption{Learning curve of e2e-Dutch on RiddleCoref dev set,
        showing performance as a function of amount of training
        data (initial segments of novels).}\label{figlearningcurve}
\end{figure}

\subsection{Results}
The main results are presented in \autoref{tblresults}.
For RiddleCoref, dutchcoref outperforms e2e-Dutch by a 6 point margin.
For SoNar-1, e2e-Dutch comes out first, and the gap is even larger.
Despite the advantage dutchcoref has due to its use of gold standard parse
trees, its performance is lower than e2e-Dutch. We can see from the mention
recall score that dutchcoref misses a large number of potential mentions; this
may be due to the fact that SoNaR-1 markables include singletons and
non-referential mentions. However, dutchcoref also has a lower LEA recall, so
the gap with e2e-Dutch on SoNar-1 is not only due to mention performance.
While results for different datasets and languages are not comparable,
the performance difference for SoNaR-1 has the same order of magnitude
as the difference for OntoNotes between the comparable rule-based and neural
systems of \citet{lee2011multipass} and \citet{lee2018higher}
in \autoref{tblenglishsota}.

RiddleCoref is much smaller than the SoNaR-1 dataset.
Is there enough training data for the neural model?
\autoref{figlearningcurve} shows a learning curve for e2e-Dutch.
This curve suggests that for the coreference scores
the answer is no, because the performance does not reach a plateau---instead
the curve is steep until the end. The performance of dutchcoref is the top of
the plot; if we extrapolate the curve linearly, we might expect
e2e-Dutch to outperform dutchcoref with 1.1--1.3
times the current training data.
However, as an anonymous reviewer pointed out,
training curves are usually logarithmic,
so more training data may be required.
Mention performance does reach a plateau,
which suggests this task is easier.

\begin{table}[t]\centering\footnotesize
\begin{tabular}{@{}llrrrrrrr@{}} 
\toprule
Novel                      & System     & \multicolumn{3}{c}{Mentions} & \multicolumn{3}{c}{LEA} & CoNLL \\ \cmidrule(lr){3-5} \cmidrule(lr){6-8}
                           &            & R     & P     & F1    & R     & P     & F1    &         \\ \midrule
Forsyth\_Cobra             & dutchcoref & 90.67 & 93.14 &\bf 91.89 & 62.82 & 74.83 &\bf 68.30 &\bf 77.42\\
Forsyth\_Cobra             & e2e-Dutch  & 78.31 & 85.23 & 81.62 & 39.82 & 44.27 & 41.93 & 55.71\\ \midrule
Japin\_Vaslav              & dutchcoref & 86.19 & 92.78 &\bf 89.36 & 44.57 & 61.39 &\bf 51.65 & 65.79\\
Japin\_Vaslav              & e2e-Dutch  & 83.13 & 91.75 & 87.22 & 49.23 & 50.89 & 50.05 &\bf 66.09\\ \midrule
Proper\_GooischeVrouwen    & dutchcoref & 88.20 & 91.12 & 89.63 & 58.65 & 66.95 &\bf 62.53 &\bf 72.29\\
Proper\_GooischeVrouwen    & e2e-Dutch  & 87.60 & 92.21 &\bf 89.85 & 50.10 & 44.74 & 47.27 & 64.77\\ \midrule
Royen\_Mannentester        & dutchcoref & 87.15 & 86.01 & 86.57 & 44.66 & 58.21 & 50.54 & 65.01\\
Royen\_Mannentester        & e2e-Dutch  & 87.90 & 89.94 &\bf 88.91 & 54.48 & 56.19 &\bf 55.32 &\bf 69.93\\ \midrule
Verhulst\_LaatsteLiefde    & dutchcoref & 86.24 & 87.70 &\bf 86.96 & 45.38 & 59.66 &\bf 51.55 &\bf 66.09\\
Verhulst\_LaatsteLiefde    & e2e-Dutch  & 82.66 & 87.98 & 85.23 & 41.58 & 48.77 & 44.89 & 61.38\\ \bottomrule
\end{tabular}
\caption{Performance difference between e2e-Dutch and dutchcoref for each
    individual novel}\label{tblresultsindiv}
\end{table}

\section{Analysis}
The previous section showed some surprising results.
We now take a closer look at the differences between the
two coreference systems, datasets, and the annotations.

\subsection{Rule-based versus neural coreference}
See \autoref{tblresultsindiv} for a novel by novel comparison
of dutchcoref and e2e-Dutch. On 3 out of 5 novels, dutchcoref is
better on both LEA F1 and CoNLL. Interestingly, on 1 novel,
LEA F1 and CoNLL disagree on the ranking of the systems.
Mention performance is high across all novels, except for a large
discrepancy on Forsyth in which e2e-Dutch scores 10 points lower.

\begin{table}[t]\centering\footnotesize
\begin{tabular}{llrrrrrrr}\toprule
System     & Dataset & Span  & Conflated & Extra   & Extra  & Divided & Missing & Missing \\
           &         & Error & Entities  & Mention & Entity & Entity  & Mention & Entity \\ \midrule
dutchcoref & RiddleCoref & 73    & 476       & 130     & 96     & 587     & 379     & 154 \\
e2e-Dutch  & RiddleCoref & 47    & 321       & 101     & 36     & 420     & 511     & 369 \\ \midrule
dutchcoref & SoNaR-1 & 352 & 2432 & 2327 & 1772 & 2640 & 2469 & 1519 \\
e2e-Dutch  & SoNaR-1 & 203 & 1187 &  895 &  695 & 1994 & 3428 & 2330 \\
\bottomrule\end{tabular}
\caption{Error types and their respective counts for both systems and datasets}\label{tblerrortypes}
\end{table}

\begin{table}[t]\centering\footnotesize
\begin{tabular}{lllrrr}\toprule
Dataset     & System    & error   &  name &  nom.\ &  pron.\ \\ \midrule
RiddleCoref & d.c.\     & extra   &     5 &       83 &       42 \\
RiddleCoref & e2e       & extra   &     6 &       55 &       40 \\
RiddleCoref & d.c.\     & missing &    11 &      163 &      205 \\
RiddleCoref & e2e       & missing &   115 &      274 &      122 \\
\midrule
SoNaR-1     & d.c.\     & extra   &   544 &     1473 &      310 \\
SoNaR-1     & e2e       & extra   &   175 &      550 &      170 \\
SoNaR-1     & d.c.\     & missing &   283 &     1842 &      344 \\
SoNaR-1     & e2e       & missing &   825 &     2124 &      479 \\
\bottomrule
\vspace{1ex}
\end{tabular}~~~~%
\begin{tabular}{llllllrrrr}\toprule
\multicolumn{3}{c}{Incorrect part} & \multicolumn{3}{c}{Rest of entity} & \multicolumn{2}{c}{Divided} & \multicolumn{2}{c}{Conflated} \\
Na  & No         & Pr         & Na          & No         & Pr         &      d.c.\ &        e2e & d.c.\ & e2e \\ 
\cmidrule(lr){1-3} \cmidrule(lr){4-6} \cmidrule(lr){7-8} \cmidrule(lr){9-10}                                      
  - &          - &         1+ &           - &         1+ &         1+ &        104 &         66 &  118 &    74 \\ 
  - &          - &         1+ &          1+ &         1+ &         1+ &        202 &         49 &   11 &    72 \\ 
  - &          - &         1+ &           - &         1+ &          - &         62 &         66 &  156 &    31 \\ 
  - &         1+ &          - &           - &         1+ &          - &         22 &         30 &   33 &    20 \\ 
  - &         1+ &         1+ &           - &         1+ &         1+ &         33 &         31 &   16 &    13 \\ 
  - &         1+ &          - &           - &         1+ &         1+ &         34 &         18 &   33 &     6 \\ 
  - &         1+ &         1+ &          1+ &         1+ &         1+ &         36 &         29 &    2 &    12 \\ 
  - &          - &         1+ &           - &          - &         1+ &         15 &         11 &   25 &    14 \\ 
\multicolumn{6}{l}{Other}                                             &         79 &        120 &   82 &    79 \\ 
\bottomrule
\end{tabular}
\caption{Left: Counts of missing and extra mention errors by mention type.
    Right: A breakdown of conflated/divided entity errors on RiddleCoref grouped by Name/Nominal/Pronoun composition;
    1+ means that the entity contains one or more mentions of the given type.
    }\label{tblmissingextramentions}\label{tblconfldivcomp}
\end{table}

To get more insight in the particular errors made by the
systems, we perform an error analysis using the tool by
\citet{kummerfeld2013error}.\footnote{%
    We adapted this tool for Dutch:
    \url{https://github.com/andreasvc/berkeley-coreference-analyser}}
This tool attributes errors to mention spans, missing or extra
mentions/entities, and entities which are divided (incorrectly split) or
conflated (incorrectly merged). We use the default configuration of ignoring
singletons mentions, but add an option to support the Dutch parse tree labels.
\autoref{tblerrortypes} shows an overview of these error types by the systems
on the RiddleCoref and SoNaR-1 test sets.
We can see that e2e-Dutch makes less errors of all types, except for missing
mentions and entities, which is due to its lower mention recall.
Even though e2e-Dutch showed a high score for mention recall on SoNaR-1 in
\autoref{tblresults}, we actually find that dutchcoref and e2e-Dutch both show
a similarly low mention recall when singletons are excluded (65.8 and 64.3,
respectively). Finally, note that a lower mention recall means that there is
less opportunity to make errors of other types, so this comparison is not
conclusive.

To understand what is going on with mention identification,
we can look at a breakdown by mention type,
see \autoref{tblmissingextramentions}. We see that e2e-Dutch produces
substantially less extra nominal (NP) mentions, but is otherwise
similar. In terms of missing mentions, e2e-Dutch makes substantially
more errors on names and nominals, but on RiddleCoref it has less missing
pronouns, while it has more missing pronouns with SoNaR-1. Although pronouns
form a closed class, the issue of pleonastic pronouns still makes pronoun
mention detection non-trivial for RiddleCoref, where pleonastic pronouns are
not annotated as mentions.
Since dutchcoref has no rules to detect non-pleonastic uses of
potentially pleonastic pronouns, it defaults to treating them as 
non-mentions. For SoNaR-1, the performance difference on missing mentions
may be due to information from the gold parse trees which is used by
dutchcoref; for example the possessive \emph{zijn} (his) has the same form as
the infinitive of the verb to be, but POS tags disambiguate this, and this
information is not available to e2e-Dutch.

Finally, we can try to understand the coreference link errors.
\autoref{tblconfldivcomp} shows the counts of link errors on RiddleCoref
by the two systems, with the entities categorized by their configuration.
We see that for both dutchcoref and e2e-Dutch,
the most common divided and conflated entity errors have a pronoun present
in the incorrect part, although dutchcoref makes more of these errors.
We can thus reconfirm the finding by \citet{kummerfeld2013error}
and \citet{vancranenburgh2019coref} who report that the most common link
error involves pronouns. Coreference resolution for Dutch provides an extra
challenge in the fact that the third person singular pronouns can refer to
either biological or linguistic gender~\citep{hoste2005optimization}.

\begin{figure}[t]\centering
    \includegraphics[width=.9\linewidth]{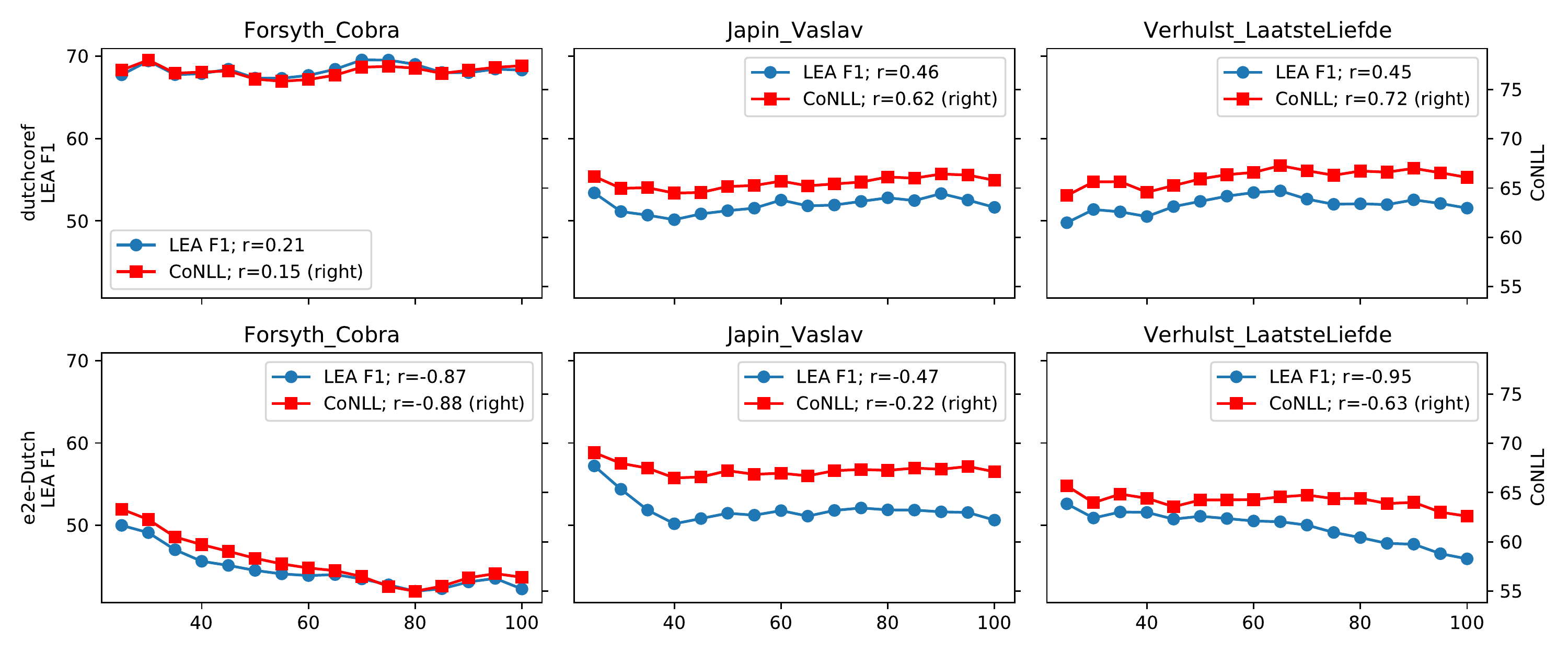}
    \caption{Coreference scores as a function of document length.
    Gold and system output are truncated at different lengths
    (based on \% of words, rounded to the nearest sentence boundary);
    $r$ is the Pearson correlation coefficient.
    }\label{figlengthcorr}
\end{figure}

\subsection{RiddleCoref (novels) versus SoNaR-1 (news/Wikipedia)}

Are the scores on the two datasets comparable?
There are several issues which hinder the comparison:
document length, domain differences, and mention annotation.

We first look at document length. It could be that the evaluation metrics
are influenced by document length, since longer documents offer
more opportunities for errors.
We will investigate this effect by truncating the documents before
evaluation, while keeping other factors such as the model or training data
constant. We truncate after running the coreference system because we want to
focus on the effect of document length on the evaluation, and we have
no reason to expect the coreference systems to behave differently on
truncated texts.
We truncate the novels at different lengths based
on the number of words, rounded to the nearest sentence.
Note that truncating does not cause additional errors, because
gold and system output are both truncated.
\autoref{figlengthcorr} shows coreference scores
as a function of document length for the novels.
We conclude that e2e-Dutch seems to perform worse
on longer documents, based on the negative correlation
of scores and document length.
While LEA weighs larger entities more, we also see
this effect with the CoNLL score, so it is not
an artifact of the LEA metric.
Moreover, we do not see the effect for dutchcoref,
so the effect is not inherent to the coreference metrics.
The documents in SoNaR-1 are much shorter
(number of sentences and words), and this may
be an advantage for e2e-Dutch.
\citet{joshi2019bert} report a similar document length effect
for English with their end-to-end model.

\autoref{tbldata} shows there is large difference in distribution of pronouns,
names, and noun phrases, which are not equally difficult.
Novels tend to have a larger proportion of pronouns.
However, it is hard to say a priori whether this would make novels easier or
more difficult in terms of coreference.

\begin{table}[t]\centering\footnotesize
\begin{tabular}{llllrrr}\toprule
System     & Dataset & Mentions & Singletons & Mentions F1 & LEA F1 & CoNLL \\ \midrule
dutchcoref & RiddleCoref  & predicted & excluded  &      80.56  & 48.15  & 56.21 \\
e2e-Dutch  & RiddleCoref  & predicted & excluded  &      79.94  & 45.31  & 54.90 \\
dutchcoref & RiddleCoref  & predicted & included  &      86.34  & 53.24  & 65.91 \\
e2e-Dutch  & RiddleCoref  & predicted & included  &      85.33  & 49.65  & 64.81 \\
dutchcoref & RiddleCoref  & gold      & included  &      100    & 61.89  & 75.84 \\
e2e-Dutch  & RiddleCoref  & gold      & included  &      100    & 55.17  & 72.01 \\
\midrule
dutchcoref & SoNaR-1   & predicted & excluded  &      63.57  & 39.71  & 46.96 \\
e2e-Dutch  & SoNaR-1   & predicted & excluded  &      67.08  & 46.18  & 52.76 \\
dutchcoref & SoNaR-1   & predicted & included  &      74.25  & 43.98  & 55.45 \\
e2e-Dutch  & SoNaR-1   & predicted & included  &      89.16  & 65.29  & 71.53 \\
dutchcoref & SoNaR-1   & gold      & included  &      100    & 59.34  & 70.90 \\
e2e-Dutch  & SoNaR-1   & gold      & included  &      100    & 74.88  & 80.61 \\ \bottomrule
\end{tabular}
\caption{Development set results under different conditions.}\label{tbldevresults}
\end{table}

In order to see the influence of the mention identification effect,
as well as the influence of evaluating with and without singletons,
\autoref{tbldevresults} shows a comparison on the development set.
Note that in our experiments with e2e-Dutch, singletons are always included
during training; excluding singletons only refers to excluding them from the
system output and gold data during evaluation. We see that ignoring singletons
has a counter-intuitively large effect on coreference scores, while it has a
relatively small effect on mention identification for RiddleCoref, but a large
effect with SoNaR-1. However, whether singletons are included or not does not
change the ranking of the systems. Finally, when gold mentions are given during
evaluation we see the large effect that mention identification has downstream,
although again the ranking is preserved.

\subsection{SoNaR-1 annotation issues}
Since the gap between the performance of e2e-Dutch and dutchcoref on SoNaR-1 is
so large, we take a quick look at the SoNaR-1 annotations of a single
development set document (WR-P-E-C-0000000021), in order to understand the
errors made by dutchcoref. However, it is apparent that
part of these errors are actually errors in the annotation.
The first thing that stands out are mentions with exact string
matches which are not linked; for example: Amsterdam (5x), Hilversum (6x),
\emph{de zeventiende eeuw} (the seventeenth century, 4x), etc. Other errors are
due to missing mentions; for example, 2 out of 10 mentions of the artist Japix
are missing, probably because the name occurs twice as part of a possessive. A
corpus based on semi-automatic annotation would not contain such errors, while
it is understandable that such links are easy to overlook in a longer document
when manually annotating from scratch.

An example of a questionable mention boundary (with corrected boundary underlined):

\ex. {[}Hij{]} was [\underline{burgemeester van Franeker} en later \underline{gedeputeerde van Friesland in de Staten-Generaal}].\\
    {[}He{]} was [\underline{mayor of Franeker} and later \underline{deputy of Frisia in the Senate}].

This is actually an example of a downside of semi-automatic annotation,
at least if there is no correction, since the markable boundaries of
SoNaR-1 were automatically extracted and
could not be changed by annotators. For the RiddleCoref corpus, such boundaries
were corrected.

An example of a missing anaphoric link (second \emph{hij} was not linked):

\ex. Een vers aan [Caspar Barlaeus]$_1$ ondertekent [hij]$_2$ met `Dando petere solitus' dat wil zeggen: [hij]$_2$ schrijft poëzie in de hoop betere verzen terug te krijgen . \\
    A verse to [Caspar Barlaeus]$_1$ he$_2$ signes with `Dando petere solitus' which is to say: he$_2$ writes poetry in the hope to get better verses back.

This only scratches the surface of the SoNaR-1 annotations. A more systematic
study should be done.

\section{Conclusion}
We found large gaps in performance for the two systems across the two domains,
but this result is not conclusive due to several reasons,
which are as follows.
The neural system shows a weakness with the long documents in the novel corpus,
but also needs more training data to reach its full potential.
The rule-based system should be better adapted to
the SoNaR-1 annotation scheme,
but the neural system's capacity to adapt to arbitrary annotation
conventions does not necessarily imply better linguistic performance.
To maximize the comparability and usefulness of the corpora,
their annotations should be harmonized, which involves manual
mention annotation.
In future work we want to improve the neural system
by using genre metadata and finetuning BERT,
and the rule-based system should be extended to a hybrid system
by adding supervised classifiers.

\section*{Acknowledgements}

We are grateful to Gertjan van Noord and Peter Kleiweg
for help with preprocessing the Lassy Small treebank,
to Wietse de Vries and Malvina Nissim for comments on the evaluation,
and to three anonymous reviewers for their suggestions.

\bibliographystyle{aclnatbib}
\bibliography{crac2020}

\end{document}